\documentclass[10pt,english,american]{article}
\usepackage[T1]{fontenc}
\usepackage[latin9]{inputenc}
\usepackage{verbatim}
\usepackage{float}
\usepackage{graphicx}

\makeatletter

\floatstyle{ruled}
\newfloat{algorithm}{tbp}{loa}
\providecommand{\algorithmname}{Algorithm}
\floatname{algorithm}{\protect\algorithmname}

\usepackage{times}
\usepackage{algorithmic}
\usepackage{spconf}
\usepackage{amsmath}
\ninept

\makeatother

\usepackage{babel}
\addto\captionsenglish{\renewcommand{\algorithmname}{Algorithm}}

\begin{document}
\selectlanguage{english}%
\name{Jean-Marc Valin$^*\dag$, Fran\c cois Michaud$\dag$ and Jean Rouat$\dag$\thanks{Part of this research was funded by the Natural Sciences and Engineering Research Council, the Canada Research Chair Program and the Canadian Foundation for Innovation.}\thanks{\copyright 2006 IEEE.  Personal use of this material is permitted. Permission from IEEE must be obtained for all other uses, in any current or future media, including reprinting/republishing this material for advertising or promotional purposes, creating new collective works, for resale or redistribution to servers or lists, or reuse of any copyrighted component of this work in other works.}}\address{$^*$CSIRO ICT Centre, Australia\hspace{0.7cm}$\dag$Universit\'e de Sherbrooke, Canada\\\texttt{jmvalin@jmvalin.ca,\{francois.michaud,jean.rouat\}@usherbrooke.ca}}

\selectlanguage{american}%

\title{Robust 3D Localization and Tracking of Sound Sources Using Beamforming
and Particle Filtering}
\maketitle
\begin{abstract}
In this paper we present a new robust sound source localization and
tracking method using an array of eight microphones (US patent pending)
. The method uses a steered beamformer based on the reliability-weighted
phase transform (RWPHAT) along with a particle filter-based tracking
algorithm. The proposed system is able to estimate both the direction
and the distance of the sources. In a videoconferencing context, the
direction was estimated with an accuracy better than one degree while
the distance was accurate within 10\% RMS. Tracking of up to three
simultaneous moving speakers is demonstrated in a noisy environment.
\end{abstract}

\section{Introduction}

Sound source localization is defined as the determination of the coordinates
of sound sources in relation to a point in space. This can be very
useful in videoconference application, either for directing the camera
toward the person speaking, or as an input to a sound source separation
algorithm \cite{ValinICASSP2004} to improve sound quality. Sound
source tracking has been demonstrated before by using Kalman filtering
\cite{Bechler2004} and particle filtering \cite{Ward2003}. However,
this has only been experimentally demonstrated with a single sound
source at a time. Our work demonstrates that it is possible to track
multiple sound sources using particle filters by solving the source-observation
assignment problem.

The proposed sound localization and tracking system is composed of
two parts: a microphone array, a memoryless localization algorithm
based on a steered beamformer, and a particle filtering tracker. The
steered beamformer is implemented in the frequency domain and scans
the space for energy peaks. The robustness of the steered beamformer
is enhanced by the use of the reliability weighted phase transform
(RWPHAT). The result of the first localization is then processed by
a particle filter that tracks each source while also preventing false
detections. 

\begin{comment}
\begin{figure}
\center{\includegraphics[width=0.7\columnwidth,keepaspectratio]{overview.eps}}

\caption{Overview of the localization system\label{cap:Overview-of-the-system}}
\end{figure}
\end{comment}

This approach improves on an earlier work in mobile robotics \cite{ValinICRA2004}
and can estimate not only the direction, but the distance of sound
sources. Localization accuracy and tracking capabilities of the system
are reported in a videoconferencing context. In that application,
the ability to estimate the source distance is significant as it solves
the parallax problem for the case when the camera is not located at
the center of the microphone array. We use a circular array because
it is the most convenient shape for our videoconferencing application.

The paper is organized as follows. Section \ref{sec:Beamformer-Based-Sound-Localization}
describes our steered beamformer based on the RWPHAT. Section \ref{sec:Particle-Based-Tracking}
explains how tracking is performed using a particle filter. This is
followed by experimental results and a discussion in Section \ref{sec:Results}.
Section \ref{sec:Conclusion} concludes the paper and presents future
work.

\begin{comment}
System Overview

The proposed localization and tracking system, as shown in Figure
\ref{cap:Overview-of-the-system}, is composed of three parts: a microphone
array, a memoryless localization algorithm based on a steered beamformer,
and a particle filtering tracker.

The array is composed of eight omni-directional microphones arranged
into a circular array on top of a table. The shape is chosen for its
symmetry and convenience in a videoconferencing setup, although the
proposed algorithm would allow other positions. The microphone signals
are used by the beamformer that scans all possible source locations
in order to maximize the output energy. The initial localization performed
by the beamformer is then used as the input of a post-processing stage
that uses particle filtering to simultaneously track all sources and
prevent false detections. In the context of videoconference, the output
of the localization system can be used to steer a camera toward the
person speaking. It can also be used as part of a source separation
algorithm to isolate the sound coming from a single source as proposed
in \cite{ValinICASSP2004}.
\end{comment}

\section{Beamformer-Based Sound Localization}

\label{sec:Beamformer-Based-Sound-Localization}

The basic idea behind the steered beamformer approach to source localization
is to steer a beamformer in all possible locations and look for maximal
output. This can be done by maximizing the output energy of a simple
delay-and-sum beamformer.

\subsection{Reliability-Weighted Phase Transform}

It was shown in \cite{ValinICRA2004} that the output energy of an
$M$-microphone delay-and-sum beamformer can be computed as a sum
of microphone pair cross-correlations $R_{x_{m_{1}},x_{m_{2}}}\left(\tau_{m_{1}}-\tau_{m_{2}}\right)$,
plus a constant microphone energy term $K$:
\begin{equation}
E=K+2\sum_{m_{1}=0}^{M-1}\sum_{m_{2}=0}^{m_{1}-1}R_{x_{m_{1}},x_{m_{2}}}\left(\tau_{m_{1}}-\tau_{m_{2}}\right)\label{eq:energy-xcorr}
\end{equation}
where $x_{m}\left(n\right)$ is the signal from the $m^{th}$ microphone
and $\tau_{m}$ is the delay of arrival (in samples) for that microphone.
Assuming that only one sound source is present, we can see that $E$
will be maximal when the delays $\tau_{m}$ are such that the microphone
signals are in phase, and therefore add constructively. 

The cross-correlation function can be approximated in the frequency
domain. A popular variation on the cross-correlation is the phase
transform (PHAT). Some of its advantages include sharper cross-correlation
peaks and a certain level of robustness to reverberation. However,
its main drawback is that all frequency bins of the spectrum have
the contribution to the final correlation, even if the signal at some
frequencies is dominated by noise or reverberation. As an improvement
over the PHAT, we introduce the reliability-weighted phase transform
(RWPHAT) defined as:
\begin{equation}
R_{i,j}^{RWPHAT}(\tau)=\sum_{k=0}^{L-1}\frac{\zeta_{i}(k)X_{i}(k)\zeta_{j}(k)X_{j}(k)^{*}}{\left|X_{i}(k)\right|\left|X_{j}(k)\right|}e^{\jmath2\pi k\tau/L}\label{eq:TDOA_correlation_weighted}
\end{equation}
where the weights $\zeta_{i}^{n}(k)$ reflect the reliability of each
frequency component. It is defined as the Wiener filter gain:
\begin{equation}
\zeta_{i}^{n}(k)=\frac{\xi_{i}^{n}(k)}{\xi_{i}^{n}(k)+1}\label{eq:SNR-weighting}
\end{equation}
where $\xi_{i}^{n}(k)$ is an estimate of the \emph{a priori} SNR
at the $i^{th}$ microphone, at time frame $n$, for frequency $k$,
computed using the decision-directed approach proposed by Ephraim
and Malah \cite{EphraimMalah1984}. 

The noise term considered for the \emph{a priori} SNR estimation is
composed of a background noise term $\sigma_{i}^{2}(k)$ and a reverberation
term $\lambda_{i}^{n}(k)$. Background noise is estimated using the
Minima-Controlled Recursive Average (MCRA) technique \cite{CohenNonStat2001},
which adapts the noise estimate during periods of low energy. We use
a simple exponential decay model for the reverberation:
\begin{equation}
\lambda_{i}^{n}(k)=\gamma\lambda_{i}^{n-1}(k)+(1-\gamma)\delta^{-1}\left|\zeta_{i}^{n}(k)X_{i}^{n-1}(k)\right|^{2}\label{eq:Reverb_weighting}
\end{equation}
where $\gamma$ is the reverberation decay (derived from the reverberation
time) of the room, $\delta$ is the signal-to-reverberant ratio (SRR)
and $R_{i}^{-1}(k)=0$. Equation \ref{eq:Reverb_weighting} can be
seen as modeling the \emph{precedence effect} \cite{Huang1999Echo},
ignoring frequency bins where a loud sound was recently present.

\subsection{Search Procedure\label{sub:Direction-Search}}

Unlike previous work using spherical mesh composed of triangular \cite{ValinICRA2004},
we now use a square grid folded onto a hemisphere. The square grid
makes it easier to do refining steps and only a hemisphere is needed
because of the ambiguity introduced by having all microphones in the
same plane. For grid parameters $u$ and $v$ in the $[-1,1]$ range,
the unit vector $\mathbf{u}$ defining the direction is expressed
as:
\begin{equation}
\mathbf{u}=\left[\frac{v}{\sqrt{u^{2}+v^{2}}}\sin\phi,\:\frac{u}{\sqrt{u^{2}+v^{2}}}\sin\phi,\:\cos\phi\right]^{\mathrm{T}}\label{eq:grid-folding}
\end{equation}
where $\phi=\pi\max\left(u^{2},v^{2}\right)/2$. The complete search
grid is defined as the space covered by $d\mathbf{u}$, where $d$
is the distance to the center of the array. 

The search for the location maximizing beamformer energy is performed
using a coarse/fine strategy. Unlike work presented by \cite{Duraiswami2001},
even the coarse search can proceed with a high resolution, with a
41x41 grid (4-degree interval) for direction and 5 possible distances.
The fine search is then used to obtain an even more accurate estimation,
with a 201x201 grid (0.9-degree interval) for direction and 25 possible
distances ranging from 30 cm to 3 meters.

The cross-correlations $R_{i,j}^{RWPHAT}(\tau)$ are computed by averaging
the cross-power spectra $X_{i}(k)X_{j}(k)^{*}$ over a time period
of 4 frames (40 ms) for overlapping windows of 1024 samples at 48
kHz. Once the cross-correlations $R_{i,j}^{RWPHAT}\!(\tau)$ are computed,
the search for the best location on the grid is performed using a
lookup-and-sum algorithm where the time delay of arrival $\tau$ for
each microphone pair and for each source location is obtained from
a lookup table. For an array of 28 microphones, this means only 28
lookup-and-sum operations for each position searched, much less than
would be required by a time-domain implementation. In the proposed
configuration ($N=8405$, $M=8$), the lookup table for the coarse
grid fits entirely in a modern processor's L2 cache, so that the algorithm
is not limited by memory access time.

After finding the loudest source by maximizing the energy of a steered
beamformer, other sources can be localized by removing the contribution
of the first source from the cross-correlations and repeating the
process. In order to remove the contribution of a source, all values
of $R_{i,j}^{RWPHAT}(\tau)$ that have been used in the sum that produced
the maximal energy are reset to zero. The process is summarized in
Algorithm \ref{cap:Steered-beamformer-direction}. Since the beamformer
does not know how many sources are present, it always looks for two
sources. This situation leads to a high rate of false detection, even
when two or more sources are present. That problem is handled by the
particle filter described in the next section.

\begin{algorithm}
\begin{algorithmic}

\FOR{$q=1$ to assumed number of sources}

\FORALL{grid index $k$}

%\STATE $E_k \leftarrow 0$

%\FORALL{microphone pair $ij$}

%\STATE $\tau \leftarrow lookup(k,ij)$

%\STATE $E_k \leftarrow E_d + R^{RWPHAT}_{i,j}(\tau)$

%\ENDFOR

\STATE $E_k \leftarrow \sum_{i,j} R^{RWPHAT}_{i,j}(lookup(k,i,j))$

\ENDFOR

\STATE $D_q \leftarrow \textrm{argmax}_k\ (E_k)$

\FORALL{microphone pair $i,j$}

%\STATE $\tau \leftarrow lookup(D_q,i,j)$

%\STATE $R^{RWPHAT}_{i,j}(\tau) \leftarrow 0$

\STATE $R^{RWPHAT}_{i,j}(lookup(D_q,i,j)) \leftarrow 0$

\ENDFOR

\ENDFOR

\end{algorithmic}

\caption{Steered beamformer location search\label{cap:Steered-beamformer-direction}}
\end{algorithm}

\section{Particle-Based Tracking}

\label{sec:Particle-Based-Tracking}

To remove false detection produced by the steered beamformer and track
each sound source, we use a probabilistic temporal integration based
on all measurements available up to the current time. It has been
shown in \cite{Ward2003,Asoh2004} that particle filters are an effective
way of tracking sound sources. Using this approach, the pdf representing
the location of each source is represented as a set of particles to
which different weights (probabilities) are assigned. The choice of
particle filtering over Kalman filtering is further justified by the
non-gaussian probabilities arising from false detections and multiple
sources.

At time $t$, we consider the case of $N_{s}$ sources ($j$ index)
being tracked, each modeled using $N_{p}$ particles ($i$ index)
of location $\mathbf{x}_{j,i}^{(t)}$ and weights $w_{j,i}^{(t)}$.
The state vector for the particles is composed of six dimensions,
three for position and three for its derivative:
\begin{equation}
\mathbf{s}_{j,i}^{(t)}=\left[\begin{array}{cc}
\mathbf{x}_{j,i}^{(t)} & \mathbf{\dot{x}}_{j,i}^{(t)}\end{array}\right]^{\mathrm{T}}\label{eq:particle_state}
\end{equation}
We implement the sampling importance resampling (SIR) algorithm. The
steps are described in the following subsections and generalize sound
source tracking to an arbitrary and non-constant number of sources. 

\begin{comment}
\begin{algorithm}
\begin{algorithmic}

\STATE 1. Predict the state $\mathbf{s}_j^{(t)}$ from $\mathbf{s}_j^{(t-1)}$ for each source $j$

\STATE 2. Compute instantaneous location probabilities from beamformer response

\STATE 3. Compute probabilities $P^{(t)}_{q,j}$ associating beamformer peaks to sources being tracked

\STATE 4. Compute updated particle weights $w^{(t)}_{j,i}$

\STATE 5. Add or remove sources if necessary

\STATE 6. Compute source localization estimates $\hat{x}_j^{(t)}$

\STATE 7. Resample particles for each source

\STATE 8. Goto 1.

\end{algorithmic}

\caption{Particle-based tracking algorithm.}
\end{algorithm}

The particle filtering algorithm is outlined in Figure \ref{cap:Particle-based-tracking-algorithm}
and generalize sound source tracking to an arbitrary and non-constant
number of sources. 

The probability density function (pdf) for the location of each source
is approximated by a set of particles that are given different weights.
The weights are updated by taking into account observations obtained
from the steered beamformer and by computing the assignment between
these observations and the sources being tracked. From there, the
estimated location of the source is the weighted mean of the particle
positions.
\end{comment}

\subsection*{Prediction}

As a predictor, we use the excitation-damping model as proposed in
\cite{Ward2003}:
\begin{eqnarray}
\mathbf{\dot{x}}_{j,i}^{(t)} & = & a\mathbf{\dot{x}}_{j,i}^{(t-1)}+bF_{\mathbf{x}}\label{eq:predict_speed}\\
\mathbf{x}_{j,i}^{(t)} & = & \mathbf{x}_{j,i}^{(t-1)}+\Delta T\mathbf{\dot{x}}_{j,i}^{(t)}\label{eq:predict_pos}
\end{eqnarray}
where $a=e^{-\alpha\Delta T}$ controls the damping term, $b=\beta\sqrt{1-a^{2}}$
controls the excitation term, $F_{\mathbf{x}}$ is a Gaussian random
variable of unit variance and $\Delta T$ is the time interval between
updates.

\subsection*{Instantaneous Location Probabilities}

The steered beamformer described in Section \ref{sec:Beamformer-Based-Sound-Localization}
produces an observation $O^{(t)}$ for each time $t$ that is composed
of $Q$ potential source locations $\mathbf{y}_{q}$. We also denote
$\mathbf{O}^{(t)}$, the set of all observations up to time $t$.
We introduce the probability $P_{q}$ that the potential source $q$
is a true source (not a false detection) that can be interpreted as
our confidence in the steered beamformer output. We know that the
higher the beamformer energy, the more likely a potential source is
to be true, so 
\begin{equation}
P_{q}=\left\{ \begin{array}{cc}
\nu^{2}/2 & \nu\leq1\\
1-\nu^{-2}/2,\qquad & \nu>1
\end{array}\right.,\;\nu=E/E_{T}\label{eq:instantaneous-prob}
\end{equation}
where $E_{T}$ is the empirical threshold energy for 50\% probability.
Assuming that $\mathbf{y}_{q}$ is not a false detection, the probability
density of observing $O_{q}^{(t)}$ for a source located at particle
position $\mathbf{x}_{j,i}^{(t)}$ is given by a normal distribution
centered at $\mathbf{x}_{j,i}$ with a standard deviation of 3 degrees
for direction and a distance-dependent standard deviation for the
distance.

\subsection*{Probabilities for Multiple Sources}

Before we can derive the update rule for the particle weights $w_{j,i}^{(t)}$,
we must first introduce the concept of source-observation assignment.
For each potential source $q$ detected by the steered beamformer,
we must compute $P_{q,j}$, the probability that the detection is
caused by the tracked source $j$, $P_{q}(H_{0})$, the probability
that the detection is a false alarm, and $P_{q}(H_{2})$, the probability
that observation $q$ corresponds to a new source.

\begin{figure}[t]
\center{\includegraphics[width=4cm,keepaspectratio]{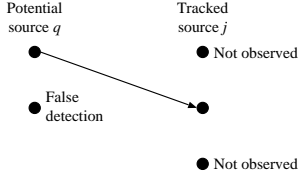}}

\caption{Assignment example where one of the tracked sources is observed and
one potential source is a false detection. The assignment can be described
as $f(\{0,1\})=\{1,-2\}$.\label{cap:Mapping-example}}
\end{figure}

Let $f:\{0,1,\ldots,Q-1\}\longrightarrow\{-2,-1,0,1,\ldots,M-1\}$
be a function assigning observations $q$ to the tracked sources $j$
(values -2 is used for false detection and -1 is used for a new source).
Figure \ref{cap:Mapping-example} illustrates a hypothetical case
with the two potential sources detected by the steered beamformer
and their assignment to the three tracked sources. Knowing $P\left(f\left|O^{(t)}\right.\right)$
(the probability that $f$ is the correct assignment given observation
$O^{(t)}$) for all possible $f$, we can compute $P_{q,j}$ as the
sum of the probabilities of all $f$ that assign potential source
$q$ to tracked source $j$. The probabilities for new sources and
false detections are obtained similarly.

Omitting $t$ for clarity, and assuming conditional independence of
the observations given the mapping function, the probability $P(f|O)$
is given by:
\begin{equation}
P(f|O)=\frac{P(f)\prod_{q}p\left(\left.O_{q}\right|f(q)\right)}{p(O)}=\frac{P(f)\prod_{q}p\left(\left.O_{q}\right|f(q)\right)}{\sum_{f}P(f)\prod_{q}p\left(\left.O_{q}\right|f(q)\right)}\label{eq:mapping-prob}
\end{equation}
We assume that the distribution of the false detections ($H_{0}$)
and the new sources ($H_{2}$) are uniform, while the distribution
for tracked sources ($H_{1}$) is the pdf approximated by the particle
distribution convolved with the steered beamformer error pdf:
\begin{equation}
p\left(\left.O_{q}\right|f(q)\right)=\sum_{i}w_{f(q),i}p\left(\left.O_{q}\right|\mathbf{x}_{j,i}\right)\label{eq:indep-inverse-mapping}
\end{equation}

The \emph{a priori} probability of $f$ being the correct assignment
is also assumed to come from independent individual components: $P(f)=\prod_{q}P\left(f(q)\right)$with:
\begin{equation}
P\left(f(q)\right)=\left\{ \begin{array}{ll}
\left(1-P_{q}\right)P_{false},\qquad & f(q)=-2\\
P_{q}P_{new} & f(q)=-1\\
P_{q}P\left(Obs_{j}^{(t)}\left|\mathbf{O}^{(t-1)}\right.\right) & f(q)\geq0
\end{array}\right.\label{eq:indep-apriori-mapping}
\end{equation}
where $P_{new}$ is the \emph{a priori} probability that a new source
appears and $P_{false}$ is the \emph{a priori} probability of false
detection and $P\left(Obs_{j}^{(t)}\left|\mathbf{O}^{(t-1)}\right.\!\right)\!=\!P\left(E_{j}\left|\mathbf{O}^{(t-1)}\right.\!\right)P\left(\mathrm{A}_{j}^{(t)}\left|\mathbf{O}^{(t-1)}\right.\!\right)$
is the probability that source $j$ is observable, i.e., that it exists
($E_{j}$) and it is active ($A_{j}^{(t)}$) at time $t$. 

The probability that the source exists is computed using Bayes law
over multiple time frames and considering the instantaneous probability
of the source being observed $P_{j}^{(t-1)}$, as well as the \emph{a~priori}
probability that the source exists despite not being observed. The
probability that a source is active (non-zero signal) is computed
by considering a first order Markov process with two states (active,
inactive). The probability that an active source remains active is
set to 0.95, and the probability that an inactive source becomes active
again is set to 0.05. We assuming that the active and inactive states
are \emph{a priori} equiprobable.

\subsection*{Weight Update}

At times $t$, assuming that the observations are conditionally independent
given the source position, and knowing that for a given source $j$,
$\sum_{i=1}^{N}w_{j,i}^{(t)}=1$, the new particle weights for source
$j$ are defined as:
\begin{equation}
w_{j,i}^{(t)}=p\left(\mathbf{x}_{j,i}^{(t)}\left|\mathbf{O}^{(t)}\right.\right)=\frac{p\left(\mathbf{x}_{j,i}^{(t)}\left|O^{(t)}\right.\right)w_{j,i}^{(t-1)}}{\sum_{i=1}^{N}p\left(\mathbf{x}_{j,i}^{(t)}\left|O^{(t)}\right.\right)w_{j,i}^{(t-1)}}\label{eq:weight_update_def}
\end{equation}
The probability $p\left(\mathbf{x}_{j,i}^{(t)}\left|O^{(t)}\right.\right)$
is given by:

\begin{equation}
p\left(\mathbf{x}_{j,i}^{(t)}\left|O^{(t)}\right.\right)=\frac{\left(1-P_{j}^{(t)}\right)}{N}+P_{j}\frac{\sum_{q}P_{q,j}^{(t)}p\left(\left.O_{q}^{(t)}\right|\mathbf{x}_{j,i}^{(t)}\right)}{\sum_{i}\sum_{q}P_{q,j}^{(t)}p\left(\left.O_{q}^{(t)}\right|\mathbf{x}_{j,i}^{(t)}\right)}\label{eq:instant-weight2}
\end{equation}

\subsection*{Adding or Removing Sources}

In a real environment, sources may appear or disappear at any moment.
If, at any time, $P_{q}(H_{2})$ is higher than a threshold equal
to $0.3$, we consider that a new source is present, in which case
a set of particles is created for source $q$. Similarly, we set a
time limit on sources so that if the source has not been observed
for a certain amount of time, we consider that it no longer exists.
In that case, the corresponding particle filter is no longer updated
nor considered in future calculations.

\subsection*{Parameter Estimation}

The estimated position of each source is the mean of the pdf and can
be obtained as a weighted average of its particles position: $\bar{\mathbf{x}}_{j}^{(t)}=\sum_{i=1}^{N}w_{j,i}^{(t)}\mathbf{x}_{j,i}^{(t)}$

\subsection*{Resampling\label{sub:Resampling}}

Resampling is performed only when $N_{eff}\approx\left(\sum_{i=1}^{N}w_{j,i}^{2}\right)^{-1}<N_{min}$
\cite{Doucet2000}. That criterion ensures that resampling only occurs
when new data is available for a certain source. Otherwise, this would
cause unnecessary reduction in particle diversity, due to some particles
randomly disappearing.

\section{Results and Discussion}

\label{sec:Results}

\begin{figure*}
\begin{center}\includegraphics[width=0.7\columnwidth,keepaspectratio]{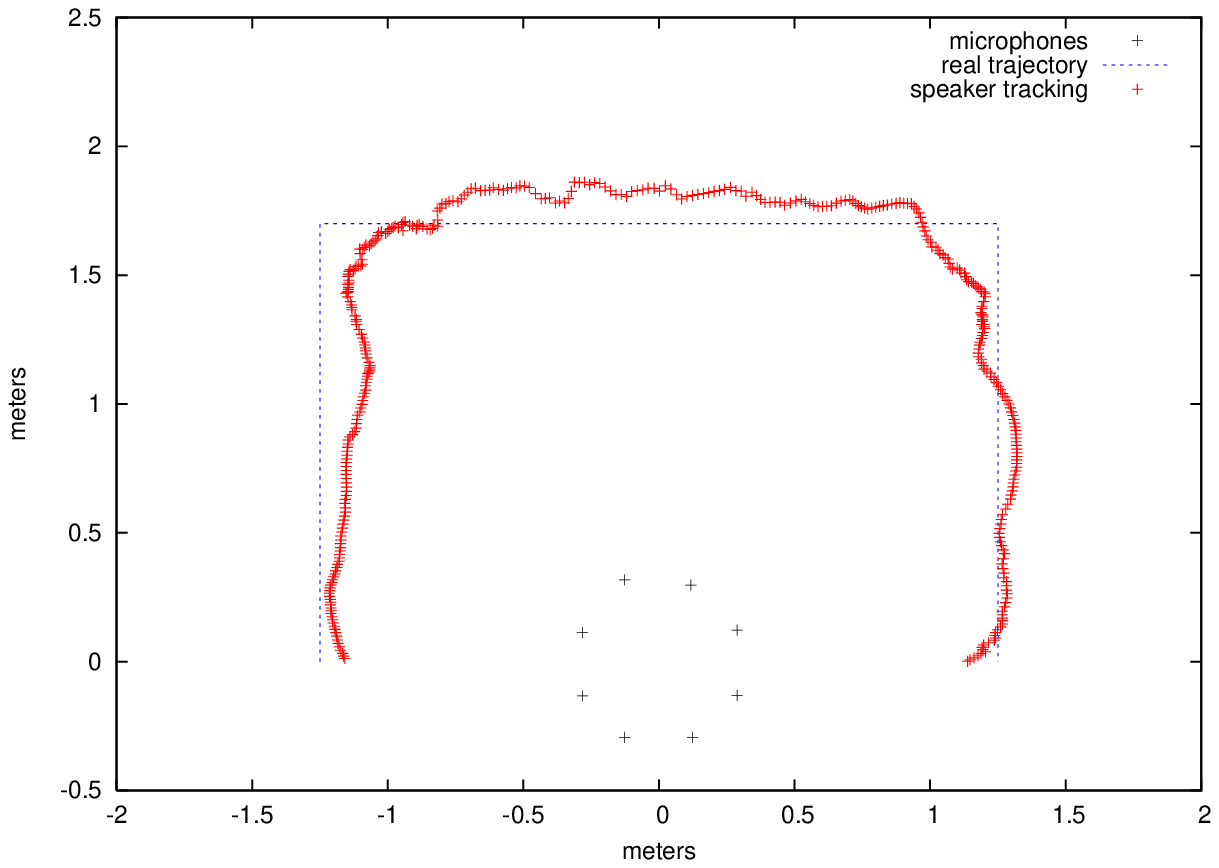}$\!\!$\includegraphics[width=0.7\columnwidth,keepaspectratio]{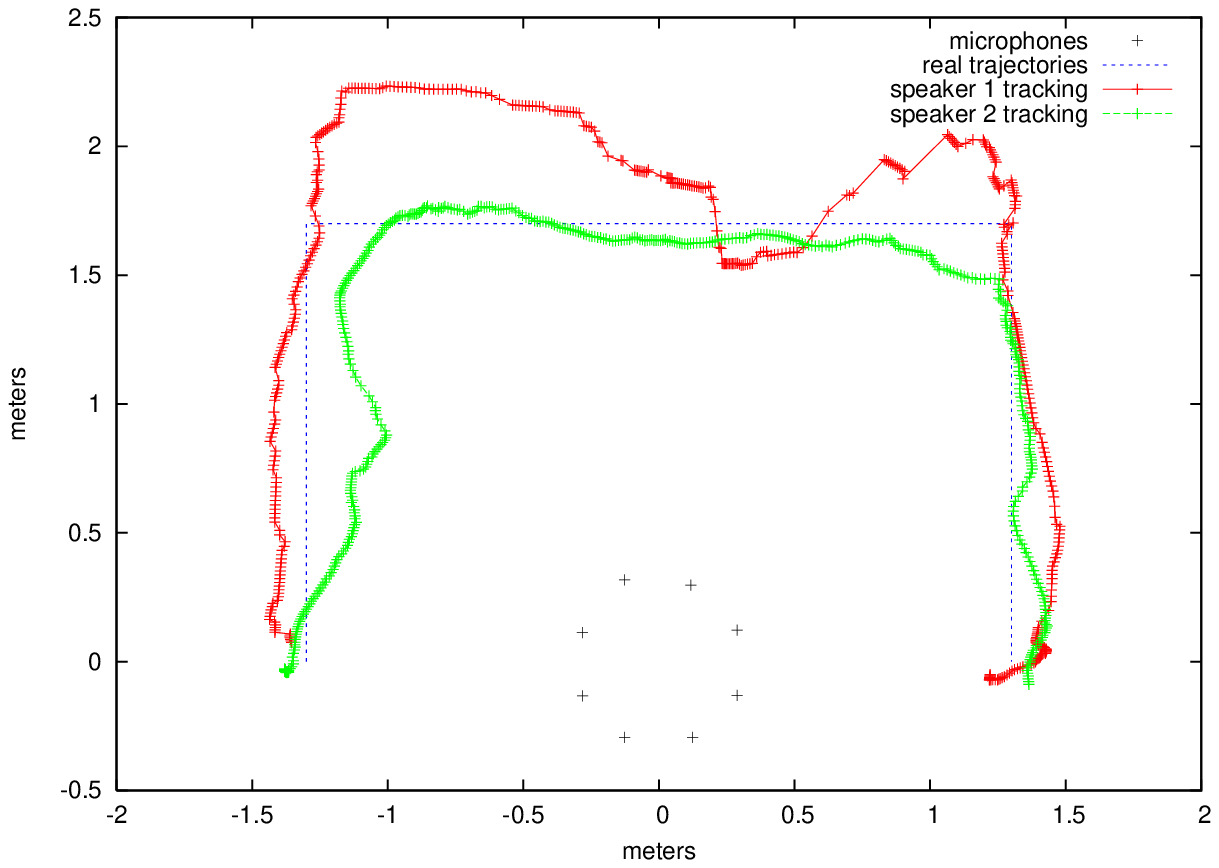}$\!\!$\includegraphics[width=0.7\columnwidth,keepaspectratio]{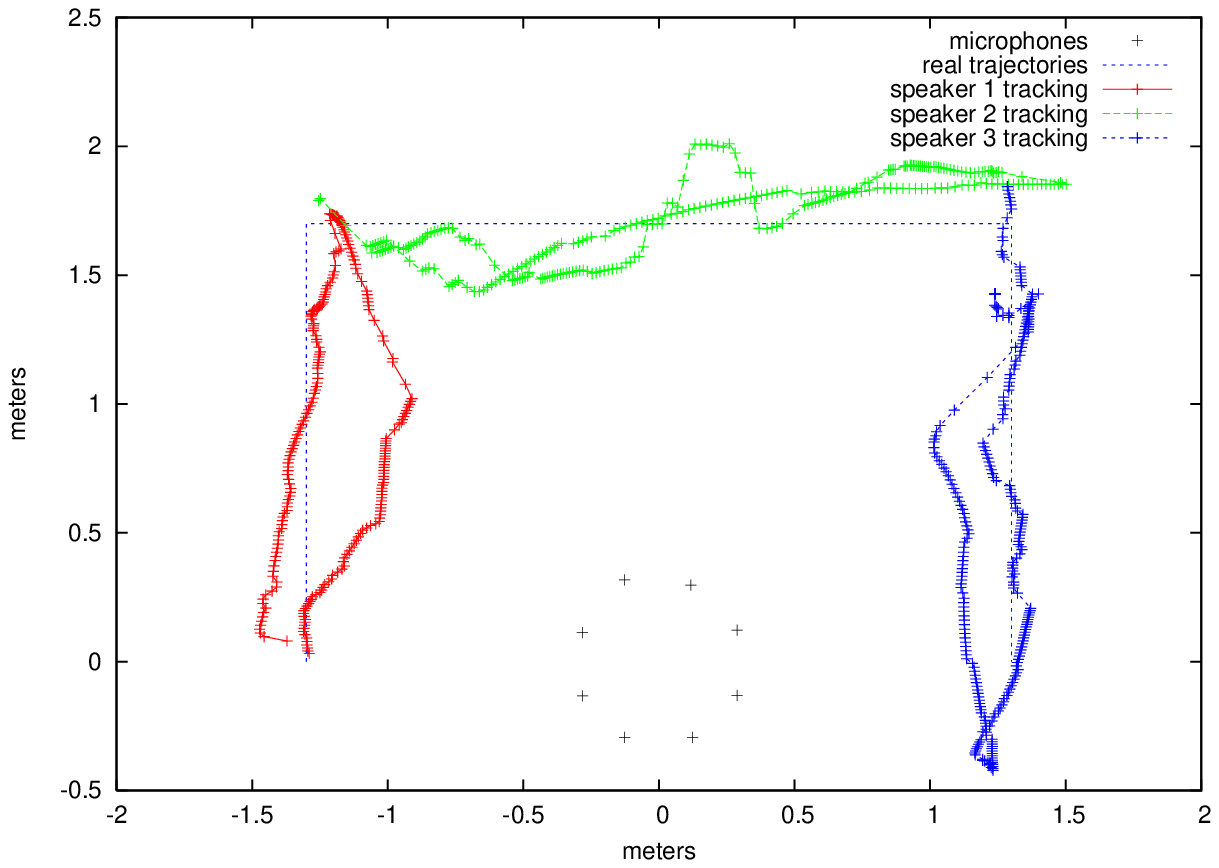}\end{center}

\caption{Tracking results in the horizontal plane (time and elevation now shown).
Left: one moving speaker (going from left to right), center: two moving
speakers (speaker 1 going from right to left, speaker 2 going from
left to right), right: three moving speakers going back and forth
on each side.\label{cap:Tracking-results}}
\end{figure*}

The proposed localization system was tested using real recordings
with a 60 cm circular array of eight omni-directional microphones
resting on top of a table. The shape of the array is chosen for its
symmetry and convenience in a videoconferencing setup, although the
proposed algorithm would allow other positions. The testing environment
is a noisy conference room resulting in an average SNR of 7 dB (assuming
one speaker) and with moderate reverberation. Running the localization
system in real-time required 30\% of a 2.13~GHz Pentium-M CPU. For
a stationary source at 1.5 meter distance, the angular accuracy was
found to be better than one degree (below our measurement accuracy)
while the distance estimate was found to have an RMS error of 10\%.
It is clear from these results that angular accuracy is much better
than distance accuracy. This is a fundamental aspect that can be explained
by the fact that distance only has a very small impact on the time
delays perceived between the microphones.

\begin{comment}
For this reason, the array we use is composed of eight omni-directional
microphones arranged into a circular array on top of a central table.
The shape is chosen for its symmetry and convenience in a videoconferencing
setup, although the proposed algorithm would allow other positions. 

Even in a case where three simultaneous stationary sources are present,
the RMS error for distance only increases slightly, to 13\%.
\end{comment}

Three tracking experiments were conducted. The results in Figure \ref{cap:Tracking-results}
show that the system is able to simultaneously track one, two or three
moving sound sources. For the case of two moving sources, the particle
filter is able to keep track of both sources even when they are crossing
in front of the array. Because we lack the ``ground truth'' position
for moving sources, only the distance error was computed\footnote{Computation uses knowledge of the height of the speakers and assumes
that the angular error is very small.} (using the information about the height of the speakers) and found
to be around 10\% for all three experiments.

\section{Conclusion}

\label{sec:Conclusion}

We have implemented a system that is able to localize and track simultaneous
moving sound sources in the presence of noise and reverberation. The
system uses an array of eight microphones and combines an RWPHAT-based
steered beamformer with a particle filter tracking algorithm capable
of following multiple sources.

An angular accuracy better than one degree was achieved with a distance
measurement error of 10\%, even for multiple moving speakers. To our
knowledge, no other work has demonstrated tracking of direction and
distance for multiple moving sound sources. The capability to track
distance is important as it will allow a camera to follow a speaker
even if it is not located at the center of the microphone array (parallax
problem).

\begin{comment}
We implemented a system with an array of eight microphones that is
able to localize and track simultaneous moving sound sources in the
presence of noise and reverberation. An angular accuracy better than
one degree was achieved with a distance measurement error of 10\%,
even for multiple moving speakers. The tracking capabilities are the
result from combining our proposed RWPHAT-based steered beamformer
with a particle filter tracking algorithm capable of following multiple
sources.

To our knowledge, no other system has demonstrated tracking of direction
and distance for multiple moving sound sources. 
\end{comment}

\bibliographystyle{IEEEbib}
\bibliography{iros,BiblioAudible,localize}

\end{document}